\pdfoutput=1
\documentclass{article}

\usepackage{iclr2027_conference,times}
\iclrfinalcopy

\usepackage[utf8]{inputenc}
\usepackage[T1]{fontenc}
\usepackage{hyperref}
\hypersetup{pdftitle={HeiCo-FOCUS: A Clinically Grounded Dataset for Long-Context Video Understanding},
  pdfauthor={Leon Mayer, Lucas Luttner, Patrick Godau, et al.}}
\usepackage{url}
\usepackage{booktabs}
\usepackage{amsmath}
\usepackage{amssymb}
\usepackage{graphicx}

\usepackage{longtable}
\usepackage{array}

\usepackage{subcaption}
\usepackage{makecell}

\usepackage{xcolor}
\usepackage{marvosym} 

\usepackage{listings}
\usepackage{acro}
\DeclareAcronym{vqa}{
    short = VQA,
    long = visual question answering
}
\DeclareAcronym{qa}{
    short = QA,
    long = question answering
}
\DeclareAcronym{vlm}{
    short = VLM,
    long = Vision-Language Model
}

\DeclareAcronym{fps}{
    short = fps,
    long = frame per second
}

\DeclareAcronym{sota}{
    short = SOTA,
    long = state-of-the-art
}

\DeclareAcronym{api}{
    short = API,
    long = Application Programming Interface
}

\DeclareAcronym{llm}{
    short = LLM,
    long = Large Language Model
}

\DeclareAcronym{ci}{
    short = CI,
    long = confidence interval
}

\DeclareAcronym{pp}{
    short = pp,
    long = percentage points
}

\title{HeiCo-FOCUS: A Clinically Grounded Dataset for Long-Context Video Understanding}

\author{%
\begin{minipage}[t]{\textwidth}\raggedright\small
Leon~Mayer\textsuperscript{1,2,$\ast$,\Letter},
Lucas~Luttner\textsuperscript{1,2,$\ast$},
Patrick~Godau\textsuperscript{1,3,$\ast$},
Kai~Fritzsche\textsuperscript{1},
Annika~Reinke\textsuperscript{1,4},
Leonie~Boland\textsuperscript{1,5},
Jule~Brandt\textsuperscript{1,2},
Janne~Heinecke\textsuperscript{1,2},
Chloe~K.~Nobuhara\textsuperscript{6},
Niklas~Holzwarth\textsuperscript{1},
Evangelia~Christodoulou\textsuperscript{1},
Marcel~Knopp\textsuperscript{1,4,5},
Dominik~Michael\textsuperscript{1,2,3},
Pascale~Piermarco\textsuperscript{1},
Saliq~Neyaz\textsuperscript{1},
Korhan~Derin~Özarslan\textsuperscript{1},
Jakob~Hennighausen\textsuperscript{1},
Carlos~Aumente-Maestro\textsuperscript{1,4},
Tim~Rädsch\textsuperscript{1,4},
Dheeraj~Baji\textsuperscript{7},
Peter~Maximilian~Full\textsuperscript{2,8},
Finn~Aichholz\textsuperscript{1},
Justus~Veit~Erpenbeck\textsuperscript{1},
Linus~Finn~Schott\textsuperscript{1},
Bastian~Winkelhausen\textsuperscript{1},
Claas~de~Boer\textsuperscript{9,10},
Bianca~Güttner\textsuperscript{9},
Anneli~Hummel\textsuperscript{9,10},
Gregor~Just\textsuperscript{9,10},
Max~Kirchner\textsuperscript{9,10},
Chenyang~Li\textsuperscript{9,10},
Rozenn~Raffaut\textsuperscript{11,12},
Ariel~Rodriguez\textsuperscript{9,10},
Danush~Kumar~Venkatesh\textsuperscript{9},
Kevin~Wang\textsuperscript{9,13},
Jinjing~Xu\textsuperscript{9,10},
Mona~Sheikh~Zeinoddin\textsuperscript{11},
Salman~Khan\textsuperscript{14},
Thomas~M.~Pausch\textsuperscript{15,16},
Stefanie~Speidel\textsuperscript{9,13},
Danail~Stoyanov\textsuperscript{11,17},
Daniel~A.~Hashimoto\textsuperscript{7,18,$\dagger$},
Fiona~R.~Kolbinger\textsuperscript{19,20,$\dagger$},
Thomas~G.~Weiser\textsuperscript{6,$\dagger$},
Lena~Maier-Hein\textsuperscript{1,2,3,4,5,21,22,$\dagger$,\Letter}
\endgraf\vspace{0.5ex}
{\normalfont\scriptsize
\textsuperscript{1}German Cancer Research Center (DKFZ) Heidelberg, Division of Intelligent Medical Systems, Germany.
\textsuperscript{2}Medical Faculty, Heidelberg University, Germany.
\textsuperscript{3}National Center for Tumor Diseases (NCT), NCT Heidelberg, Germany.
\textsuperscript{4}Helmholtz Imaging, German Cancer Research Center (DKFZ), Germany.
\textsuperscript{5}Faculty of Mathematics and Computer Science, Heidelberg University, Heidelberg, Germany.
\textsuperscript{6}Department of Surgery, Stanford University, Stanford, CA, USA.
\textsuperscript{7}Department of Surgery, Perelman School of Medicine, University of Pennsylvania, Philadelphia, PA, USA.
\textsuperscript{8}Division of Medical Image Computing, German Cancer Research Center (DKFZ), Heidelberg, Germany.
\textsuperscript{9}Department of Translational Surgical Oncology, National Center for Tumor Diseases (NCT), NCT/UCC Dresden, a partnership between DKFZ, Faculty of Medicine and University Hospital Carl Gustav Carus, TUD Dresden University of Technology, and Helmholtz-Zentrum Dresden-Rossendorf (HZDR), Germany.
\textsuperscript{10}Department of Translational Surgical Oncology, NCT/UCC Dresden, Faculty of Medicine and University Hospital Carl Gustav Carus, TUD Dresden, Dresden, Germany.
\textsuperscript{11}UCL Hawkes Institute, University College London, London, UK.
\textsuperscript{12}Department of Medical Physics and Biomedical Engineering, University College London, London WC1E 6BT, UK.
\textsuperscript{13}Cluster of Excellence-CeTI, TUD, Germany.
\textsuperscript{14}Mohamed bin Zayed University of Artificial Intelligence, Abu Dhabi, UAE.
\textsuperscript{15}Department of General, Visceral and Transplantation Surgery, Heidelberg University Hospital, Heidelberg, Germany.
\textsuperscript{16}Institute of Medical Informatics (IMI), Heidelberg University, Heidelberg, Germany.
\textsuperscript{17}Department of Computer Science, University College London, London WC1E 6BT, UK.
\textsuperscript{18}Department of Computer and Information Science, University of Pennsylvania, Philadelphia, PA, USA.
\textsuperscript{19}Weldon School of Biomedical Engineering, Purdue University, West Lafayette, IN, USA.
\textsuperscript{20}Department of Visceral, Thoracic and Vascular Surgery, University Hospital and Faculty of Medicine Carl Gustav Carus, TUD Dresden University of Technology, Dresden, Germany.
\textsuperscript{21}HIDSS4Health -- Helmholtz Information and Data Science School for Health, Karlsruhe/Heidelberg, Germany.
\textsuperscript{22}Heidelberg University Hospital, Surgical Clinic, Surgical AI Research Group, Heidelberg, Germany.
$^{\ast}$Contributed equally. Each co-first author may list themselves as lead author on their CV.\quad
$^{\dagger}$Shared last authorship.\quad
\textsuperscript{\Letter}Corresponding authors: \href{mailto:leon.mayer@dkfz-heidelberg.de}{leon.mayer@dkfz-heidelberg.de}, \href{mailto:l.maier-hein@dkfz-heidelberg.de}{l.maier-hein@dkfz-heidelberg.de}\endgraf}
\end{minipage}}

\begin{document}

\maketitle
\lhead{Preprint}

\acresetall
\vspace{-0.15in}
\begin{abstract}
Recent advances in \acfp{vlm} have led to rapid progress in video understanding across a wide range of benchmark tasks. However, existing evaluations largely focus on short-term reasoning, failing to assess a critical capability: maintaining cumulative temporal consistency over extended time horizons. To close this evaluation gap, we introduce HeiCo-FOCUS, a clinically grounded dataset for evaluating long-context video understanding through the task of \textbf{F}oreign \textbf{O}bject \textbf{C}ontextual \textbf{U}nderstanding in \textbf{S}urgery. Built on a dataset of \textbf{Hei}delberg \textbf{Co}lorectal surgeries, this task requires models to continuously track multiple objects as they are inserted, manipulated, occluded, and removed over procedures lasting up to hours. 
HeiCo-FOCUS comprises 30,000 \ac{vqa} pairs covering five core capabilities: object recognition, temporal grounding, aggregation, event and procedural understanding, and complex reasoning. The dataset was constructed through a rigorous multi-stage annotation pipeline involving large-scale crowd annotation and 39 surgical domain experts to ensure high quality and clinical relevance. To systematically probe model behavior, we introduce a multi-track evaluation framework that progressively increases temporal and contextual demands from single frames to full procedures.
Experiments with ten frontier \acp{vlm} show that HeiCo-FOCUS tasks are far from solved: only around half of the models clearly outperform a text-only baseline. Across the video tracks, models perform best on \textit{event and procedural understanding} (mean Accuracy: 56.5\% across all models), while \textit{temporal grounding} remains particularly challenging for all evaluated models (mean Accuracy: 19.7\%). We therefore expect HeiCo-FOCUS to serve as a catalyst for the development of models capable of reliable, temporally consistent reasoning over hours-long videos.

\textit{Code is available at \url{https://github.com/IMSY-DKFZ/orena-focus}. The data will be published on Hugging Face soon; until then, please contact the corresponding authors for access.}
\end{abstract}

\acresetall

\section{Introduction}
\label{sec:intro}

\begin{figure*}[t]
    
    \centering
    \includegraphics[width=\linewidth]{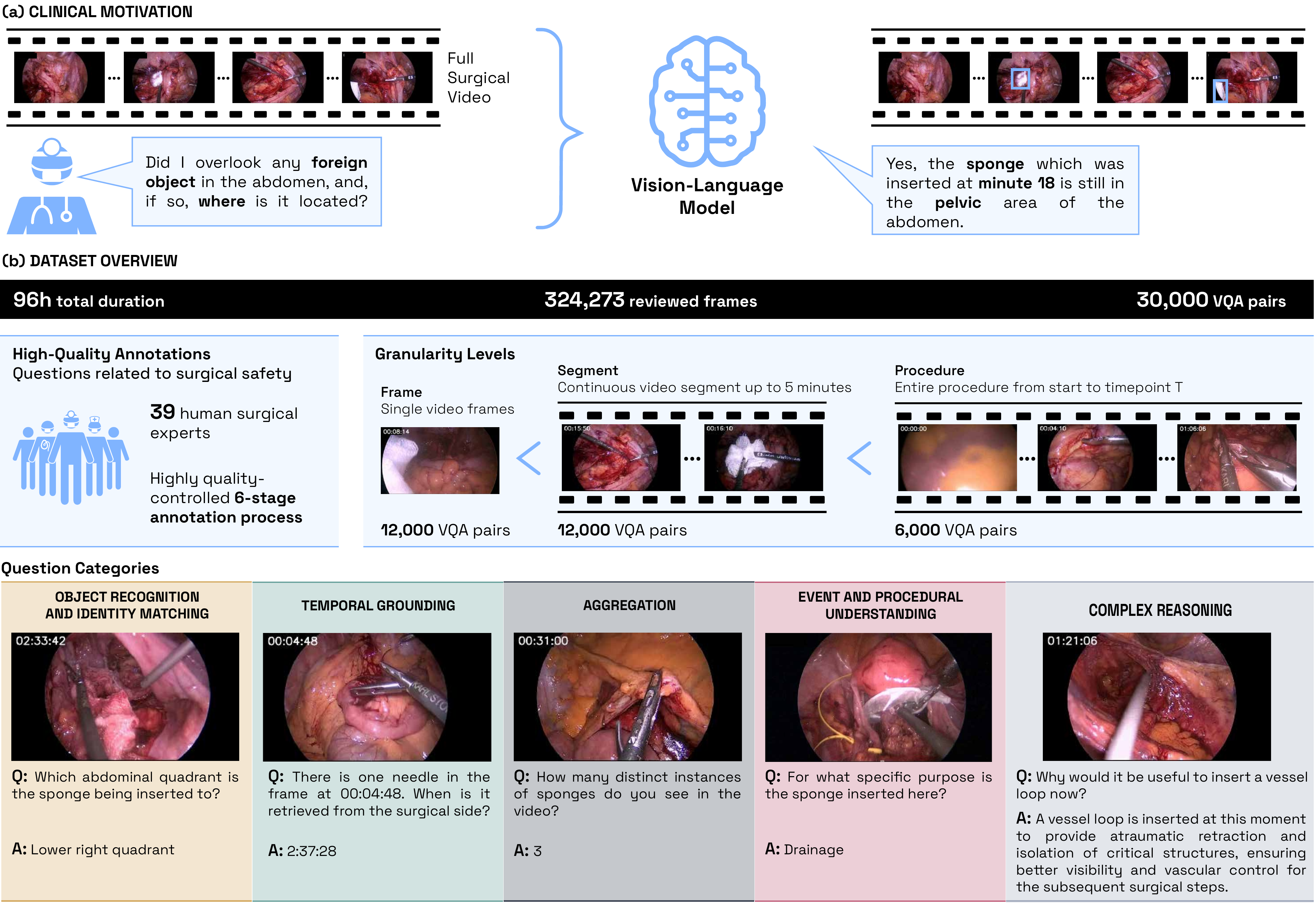}
    \caption{\textbf{Overview of the HeiCo-FOCUS benchmark. a) Clinical motivation.} Ensuring the retrieval of foreign objects at the end of a surgery is critical, as retained objects can lead to severe complications.  \textbf{b) Dataset overview.} The dataset comprises 96 hours of surgical video annotated with 30,000 \acf{vqa} pairs spanning five core capabilities (see Figure~\ref{fig:taxonomy}).}
    \label{fig:HeiCo-FOCUS-overview}
\end{figure*}
Recent progress in general-domain \acp{vlm} has enabled temporal reasoning over extended video streams \citep{bai2025qwen3, comanici2025gemini}. Emerging long-video benchmarks evaluate these capabilities, showing improvements in tasks such as \ac{vqa} and event understanding over longer temporal contexts \citep{wu2024longvideobench, chandrasegaran2024hourvideo, fu2025video, shen2024longvu, zohar2025apollo, wang2025lvbench, liu2024mmbench, zhou2025mlvu, hong2023lvos}. However, these benchmarks primarily assess local reasoning within extended segments and do not require persistent tracking of entities across long horizons or the aggregation of state over time. 

Surgical videos represent a particularly important and challenging domain: procedures unfold over tens of minutes to hours, involve continuous scene changes, and require precise interpretation of fine-grained visual cues. Despite this, the surgical AI community has so far lacked a dedicated, standardized benchmark to assess whether emerging long-context video capabilities translate to real clinical workflows (see Table~\ref{tab:surgical_vqa_comparison} in Appendix~\ref{app:dataset_comparison}). The HeiCo-FOCUS dataset addresses this unmet need by introducing a structured benchmark centered on a concrete patient safety problem: ensuring the retrieval of foreign objects at the end of surgery (see Figure~\ref{fig:HeiCo-FOCUS-overview}). In minimally invasive surgical procedures, surgical sponges, sutures, and other foreign objects are inserted into the abdomen via trocar sites to carry out steps of the procedure. Unintentionally leaving foreign objects in the abdomen is a rare but consequential incident, as they can cause serious complications \citep{badiee2025retained, Weprin2021RetainedSharps}. Using foreign objects as the central theme, HeiCo-FOCUS offers comprehensive capability coverage, spanning object recognition, temporal grounding, aggregation, event and procedural understanding, and complex reasoning, enabling fine-grained analysis of model capabilities (Figure~\ref{fig:taxonomy}). 
From a long video benchmarking perspective, keeping track of foreign objects is a highly attractive application because it inherently requires persistent memory, temporal consistency, and aggregation over extended time horizons. 

The uniqueness of HeiCo-FOCUS is rooted in three fundamental design decisions:
\begin{enumerate}
    \item \textbf{Quality-first approach:} While large-scale, large language model (LLM) curated datasets have significantly accelerated progress in AI, they primarily serve the purpose of training models and inevitably contain noise. In contrast, evaluation benchmarks require the highest possible level of reliability, as their credibility directly determines the validity of scientific comparisons. Recent work in medical AI has highlighted how insufficient data provenance can undermine research quality and even contribute to paper retractions \citep{gibson2026unreliable}. \textbf{HeiCo-FOCUS directly addresses this critical gap by providing a fully trusted benchmark:} (1) building upon a trusted, clinically curated base dataset (HeiCo; \citealp{maier2021heidelberg}), and (2) involving human experts at every stage of the annotation pipeline to ensure high-quality, clinically meaningful annotations and evaluation. Notably, compared to more than 400 analyzed medical imaging AI benchmarks (median: 3 annotators) \citep{reinke2026competitions}, HeiCo-FOCUS involved more than an order of magnitude more domain experts, reflecting the substantial annotation effort required to produce a clinically reliable benchmark.
    \item \textbf{Persistent object identities for long-context \ac{vqa}:} Existing surgical \ac{vqa} datasets largely generate question-answer pairs by converting existing vision annotations using large language models. This introduces a fundamental limitation: many clinically relevant questions are never covered. For example, questions related to retained foreign objects require persistent object identities over long time horizons. Existing datasets do not provide instance-level annotations, current models cannot reliably generate them automatically, and manual annotation is prohibitively expensive as illustrated by the enormous cost of our dataset. Furthermore, automatically generated questions are inherently limited by the underlying annotations and often fail to capture clinically meaningful long-context reasoning. HeiCo-FOCUS closes this gap through a dedicated, expert-driven annotation pipeline that explicitly annotates object instances (Figure~\ref{fig:annotation_pipeline}). \textbf{Without HeiCo-FOCUS, the community will lack a clinically grounded benchmark for systematically evaluating long-context understanding tasks that require persistent object identities.}
    \item \textbf{Clinical relevance:} Many surgical AI datasets have been designed around procedures and tasks that are easiest to acquire and annotate, rather than those with the greatest clinical need. This is nicely illustrated by a recent scoping review \citep{carstens2025surgical} that showed that laparoscopic cholecystectomy accounts for more than 50\% of surgical AI studies and has therefore largely driven progress in the field. This is interesting because cholecystectomies are technically straightforward procedures with low complication rates ($\sim$2--3\%; \citealp{duca2003laparoscopic}), suggesting that their predominance is driven by available landmark datasets. As datasets shape research priorities, benchmarks should steer research toward clinically important problems. Uniquely, HeiCo-FOCUS addresses both: a huge technical challenge (long-context video understanding) and an actual globally important clinical problem. \textbf{Instead of further expanding cholecystectomy datasets, we deliberately chose colorectal procedures (complications in up to one third of patients; \citealp{tevis2016postoperative}), which present an up to 15 times higher complication rate, 5-fold longer procedures, and on average three times more foreign object instances per video.}
\end{enumerate}

Applying \ac{sota} \acp{vlm} to HeiCo-FOCUS highlights current limitations in long-context understanding, temporal consistency, and safety-critical reasoning. Notably, HeiCo-FOCUS pushes existing models to their limits, indicating that the benchmark is unlikely to be saturated in the near term (Figure~\ref{fig:results_trackwise}).

\section{Related Work}
\label{sec:related_work}

Long video benchmarks, such as \citep{wu2024longvideobench, chandrasegaran2024hourvideo, fu2025video, shen2024longvu, zohar2025apollo, wang2025lvbench, liu2024mmbench, zhou2025mlvu, hong2023lvos, mangalam_egoschema_2023, wu2025moviebench, yu2019activitynet} have been instrumental in advancing video understanding beyond short clips by introducing tasks that require reasoning over extended temporal contexts. However, they still typically rely on relatively short temporal windows---often limited to a few minutes.%

Several groups have pioneered the development of surgical \ac{vqa} datasets, with existing resources summarized in Table~\ref{tab:surgical_vqa_comparison} (Appendix~\ref{app:dataset_comparison}). These datasets have significantly advanced the field by providing large-scale question–answer pairs and enabling the training of multimodal models in surgical domains. In particular, recent datasets have reached substantial scale in terms of both video volume and number of \ac{vqa} pairs. However, despite these advances, several important limitations remain. First, existing datasets are restricted to frame-level or short video clips, with little to no support for long-context reasoning beyond a few minutes. As a result, they do not capture the temporal dependencies and cumulative reasoning required in real surgical workflows. Second, questions are often limited to perception and event understanding. We are not aware of a single surgical AI dataset that addresses temporal grounding and long-context reasoning. Third, many datasets rely either on rule-based categorical question generation \citep{seenivasan2022surgical} or on AI-generated \ac{vqa} pairs sourced from public platforms like YouTube, utilizing automated audio transcription and rephrasing \citep{li2026surgpub, perez2026surglavi}. Fourth, expert verification is largely absent, with only a small fraction of datasets reporting systematic review by medical professionals. 

Together, these limitations highlight the need for a benchmark that combines long-context video understanding, clinically grounded question design, and rigorous expert validation.

\section{HeiCo-FOCUS}
\label{sec:FOCUS}

HeiCo-FOCUS is a clinically grounded benchmark for evaluating long-context video understanding. Based on all 30 \textbf{Hei}delberg \textbf{C}olorectal Surgeries of the HeiCo dataset \citep{maier2021heidelberg}, it contributes 30,000 \ac{vqa} pairs addressing \textbf{F}oreign \textbf{O}bject \textbf{C}ontextual \textbf{U}nderstanding in \textbf{S}urgery.

\paragraph{Dataset overview and license.} The original HeiCo dataset comprises a total of 30 colorectal surgical procedures, including 10 sigmoid resections, 10 rectal resections, and 10 proctocolectomies, annotated with surgical phases and instruments. We deliberately build upon this trusted resource to (1) leverage an established, ethically governed clinical dataset with transparent provenance, (2) focus on complex procedures with substantial complication rates and a broad diversity of foreign objects, and (3) enable researchers to combine HeiCo-FOCUS with the rich existing HeiCo annotations, facilitating complementary analyses of surgical phases, instruments, and long-context object dynamics on the same procedures. For each procedure, we generated 1,000 \ac{vqa} pairs, comprising 400 questions for the FRAME track, 400 for the SEGMENT track, and 200 for the PROCEDURE track. The \ac{vqa} generation pipeline (see Section~\ref{sec:vqa_generation_pipeline}) was designed to achieve approximately balanced coverage across all capability categories and temporal contexts. To support fine-grained evaluation and stratified analysis, each \ac{vqa} pair is associated with structured metadata including the expected answer format (e.g., open ended, an integer number, a time estimate, etc.) and associated capabilities (see Section~\ref{subsec:vqa_generation_pipeline_taxonomy}). Following the original HeiCo dataset split~\citep{maier2021heidelberg}, we recommend using the sigmoid resection procedures as the held-out test set (4,000 FRAME, 4,000 SEGMENT, and 2,000 PROCEDURE questions). To comply with the original HeiCo license, HeiCo-FOCUS is released under CC BY-NC-SA 4.0. Usage and access details are presented in Appendix~\ref{app:data_access_usage}.

The design of HeiCo-FOCUS was guided by the following principles: 
\paragraph{Clinical relevance and real-world grounding.} A central design objective was to ensure that the benchmark is both clinically meaningful and technically rigorous. To achieve this, surgical experts were closely involved throughout all stages of the benchmark development. 

\paragraph{Fine-grained understanding of \ac{vlm} capabilities.} To enable a detailed analysis of model behavior, we developed a comprehensive capability taxonomy (see Figure~\ref{fig:taxonomy}) that spans multiple dimensions of visual and temporal understanding. Secondly, to allow for controlled analysis of how model performance evolves from instantaneous perception to long-horizon reasoning, the benchmark was structured into three complementary tracks: a \textbf{FRAME Track} to assess single-image understanding, focusing on perception and aggregation tasks, a \textbf{SEGMENT Track} to assess short-term video understanding (all capabilities) within video segments of up to 5 min and a  \textbf{PROCEDURE Track} to assess long-context capabilities over extended video durations from the beginning up to a time point \textit{t} (5--296 min). 

\paragraph{Quality-controlled annotation.} To ensure high annotation quality, we implemented a multi-stage annotation pipeline (see Figure~\ref{fig:annotation_pipeline}, Section~\ref{subsec:vqa_generation_pipeline_annotation}, and Table~\ref{tab:annotation_stages} in Appendix~\ref{app:vqa_generation_process}) combining automated methods, large-scale crowd annotation, and expert annotation and verification. Overall, 39 domain experts (clinical and technical) were involved in the data annotation (7 expert surgeons, 18 medical students, 14 surgical AI researchers). 

\paragraph{Bias-aware validation.} To prevent questions from being answered based solely on prior knowledge or textual plausibility \citep{mayer20256, asadi2026mirage}, all questions were filtered with a panel of text-only models (\textit{blind filter}), as detailed below. Furthermore, the capability taxonomy enables stratified analyses across different capabilities, preventing overrepresentation of simpler tasks (e.g., object identification) from dominating performance estimates. Finally, the hierarchical structure of the data - where multiple questions are associated with the same video or segment - is explicitly accounted for in the statistical analysis, for example through hierarchical aggregation and bootstrapping.

\section{\ac{vqa} Generation Pipeline}
\label{sec:vqa_generation_pipeline}
The HeiCo-FOCUS dataset was constructed through a multi-stage annotation pipeline designed to balance scalability with high clinical and technical quality. %
\subsection{Taxonomy}
\label{subsec:vqa_generation_pipeline_taxonomy}
\begin{figure*}[t]
    \centering
    \includegraphics[width=0.8\linewidth]{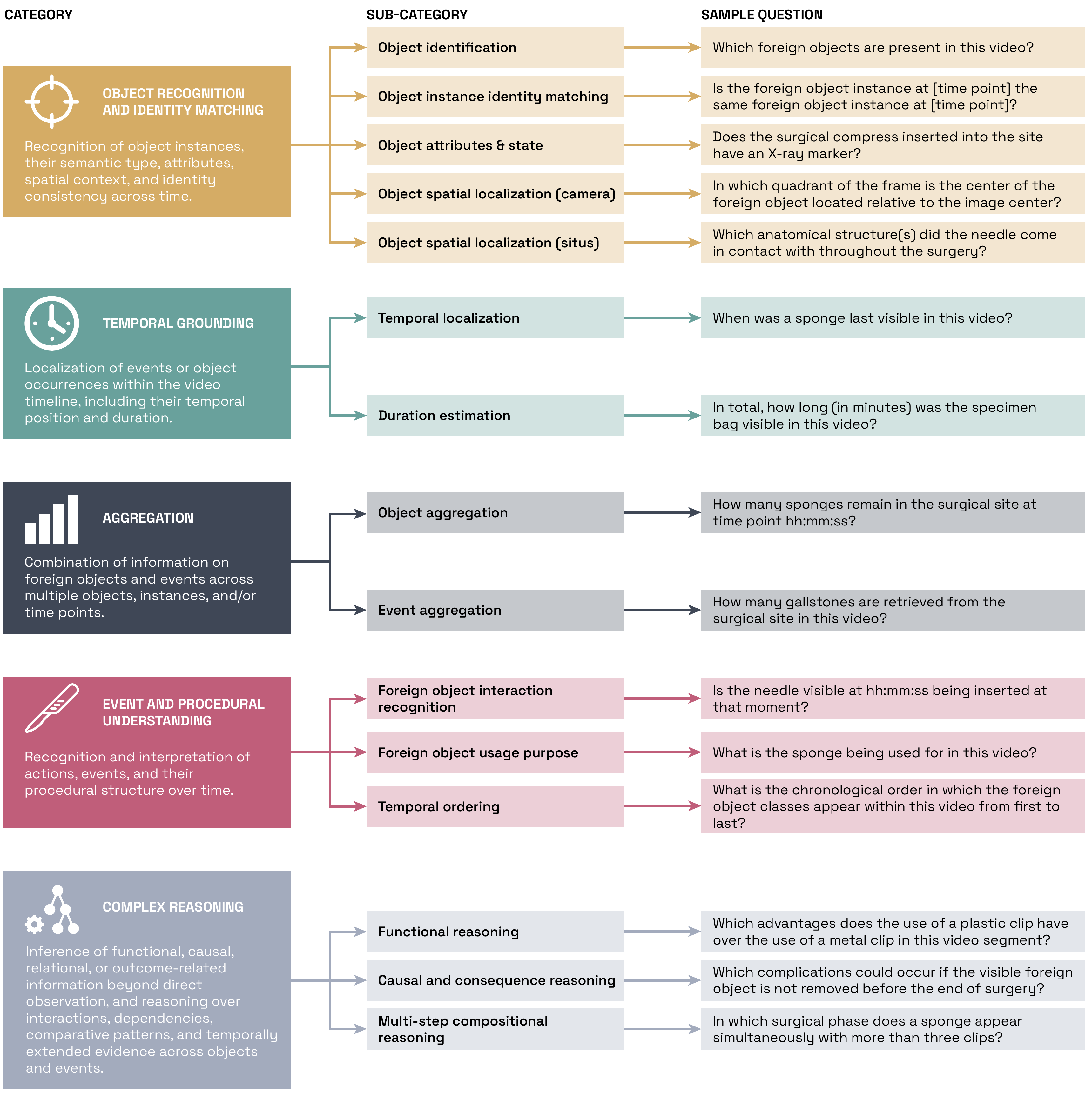}
    \caption{\textbf{HeiCo-FOCUS taxonomy with sample questions}; definitions in Appendix~\ref{app:taxonomy}.}
    \label{fig:taxonomy}
\end{figure*}

Our taxonomy for surgical \ac{vqa} is depicted in Figure~\ref{fig:taxonomy}. For question categorization, each question is assigned to one primary capability category and may optionally be tagged with additional secondary sub-categories to reflect its multi-faceted nature. The primary category is determined by the dominant reasoning requirement of the question. For example, \textit{“Where”} questions are assigned to \textit{object recognition and identity matching $\rightarrow$ spatial localization}, \textit{“How many”} questions to \textit{aggregation}, and \textit{“When”} questions to \textit{temporal grounding $\rightarrow$ temporal localization}. This primary assignment enables the balancing of \ac{vqa} pairs across capabilities while secondary labels allow fine-grained stratified evaluation. The complete taxonomy is provided in Appendix~\ref{app:taxonomy}.
\subsection{Annotation Pipeline}
\label{subsec:vqa_generation_pipeline_annotation}

Our annotation pipeline combines large-scale crowd annotation, automated processing, and expert-driven refinement to progressively enrich the data from raw video to clinically meaningful and verified \ac{vqa} pairs, as illustrated in Figure~\ref{fig:annotation_pipeline} and detailed in Appendix~\ref{app:vqa_generation_process}.

The process began with \textbf{Stage 1}, in which short video clips were screened to identify the presence of foreign objects. In \textbf{Stage 2}, frames sampled at 1 \ac{fps} from relevant segments were annotated with bounding boxes and class labels. These initial annotations were subsequently refined in \textbf{Stage 3} with automated methods that correct and complete bounding box predictions.

\textbf{Stage 4} was by far the most resource-intensive one. Here, 18 medical students assigned consistent instance identities to individual objects across time, yielding a structured representation of foreign object trajectories. This required the generation of bounding boxes around foreign objects at 1 \ac{fps} and for more than 300 k frames in total. Appendix~\ref{app:challenges} illustrates annotation challenges.

Building on the fine-grained instance-level annotations, \textbf{Stage 5} derived anchor moments, such as insertion and retrieval events, long absences, and cluttered scenes, which served as key reference points for \ac{vqa} generation. \ac{vqa} pairs were then created through a combination of automated template-based methods filtered by medical students for validity, and surgical expert-designed questions, ensuring coverage across capability categories and complexity levels. Finally, in \textbf{Stage 6}, the whole \ac{vqa} generation pipeline was verified by the analysis of about 500 diverse sample questions, covering all capabilities and foreign objects. 

Once the complete annotation pipeline had been validated, it was applied to generate a total of 30,000 \ac{vqa} pairs for the HeiCo-FOCUS benchmark, as detailed in the following section. To further promote transparency, quality control, and continuous improvement, we additionally provide a community feedback mechanism that enables users to report potentially ambiguous or erroneous cases during a restricted post-release review period.

\begin{figure*}
    \centering
    \includegraphics[width=0.9\linewidth]{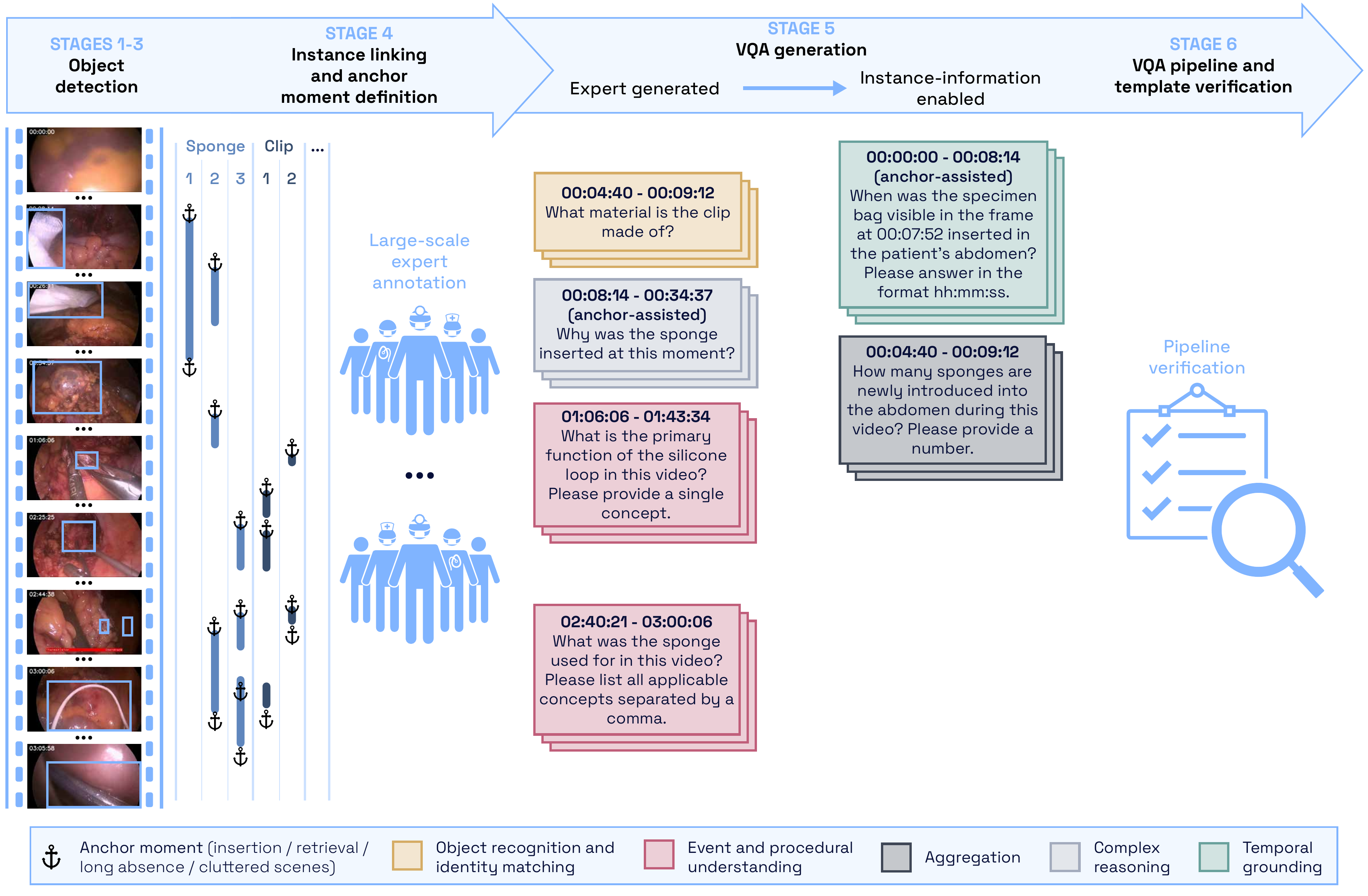}
    \caption{\textbf{Multi-stage annotation pipeline of HeiCo-FOCUS.} Raw surgical videos are progressively transformed into high-quality, clinically grounded \acf{vqa} pairs. Stages 1-3 perform foreign object detection, followed by Stage 4, where instances are linked across time. In Stage 5, \ac{vqa} pairs are generated using a combination of expert-designed questions and instance-informed automatic templates. Stage 6 conducts end-to-end expert verification.}
    \label{fig:annotation_pipeline}
\end{figure*}

\subsection{\ac{vqa} Design}
\label{subsec:vqa_generation_pipeline_vqa_design}

The \ac{vqa} generation process builds upon Stage 4 of the annotation pipeline (see Figure~\ref{fig:annotation_pipeline}), which resulted in 324,273 manually curated frames containing 79,682 verified foreign-object bounding boxes. Every class except clips carries instance identities across time (55,629 boxes); as it turned out to be infeasible to do instance linking for clips (even for expert surgeons), they were annotated for visibility only and excluded from questions that require following an individual object over time. These high-quality instance-level annotations provide a structured representation of object trajectories over time and enable automatic generation of questions for a large subset of capabilities, including temporal grounding, aggregation, and multi-step compositional reasoning. However, several sub-capabilities-particularly functional reasoning, causal reasoning, and anatomically grounded localization-required additional clinical expertise beyond what can be derived from instance annotations alone. To ensure both high clinical validity and broad capability coverage, the question generation therefore followed a multi-stage, expert-guided pipeline combining manual annotation, expert validation, and automated augmentation. 

\paragraph{Candidate question and anchor moment generation.} We first defined a catalog of candidate question templates covering all benchmark tracks and capability categories (e.g., “How many <foreign object>s were inserted during this video segment”). Each template was associated with metadata specifying its applicable track, primary and secondary capability labels, relevant foreign object classes, required visual evidence, and answer format. In parallel, we derived informative anchor moments from the instance-level annotations, including potential insertion and retrieval events, long absence intervals, and visually cluttered frames or segments. These anchors served as high-value time points for clinically meaningful, technically challenging questions.

\paragraph{Expert annotations.} Based on question and anchor moment candidates, medical students validated automatically generated anchor moments, confirming insertion and retrieval events and pre-filtered generated questions. These questions were either answered when confidence was high, discarded if ill-posed, or flagged for expert review when uncertainty remained. Next, expert surgeons reviewed these flagged cases, validated anchor moments, and provided authoritative answers. In addition, they enriched the dataset by contributing new questions that could be asked at any time during the procedures. In this step, experts were asked to target underrepresented capabilities such as reasoning, functional understanding, and anatomically grounded localization, for which questions cannot be automatically answered based on instance annotations.

\paragraph{Template-based question generation and balancing.} To achieve a balanced benchmark, clinician-written questions were complemented with question templates systematically instantiated both at clinically informative anchor moments and at other eligible time points to ensure broad coverage across capabilities, foreign-object classes, answer values, and temporal contexts. For sampling, the taxonomy was represented as a tree whose leaves contained 85 question templates, and questions were iteratively drawn from the least represented capability, sub-capability, template, and template instance. Within templates, sampling was additionally balanced across answer values and distributed over time.

After end-to-end verification of the complete annotation process, we followed the described steps to generate 1,000 \ac{vqa} pairs per video. To remove questions answerable from text alone, three models (Mistral Small 2603, Gemini 2.5 Flash Lite, and DeepSeek V4 Flash) answered every candidate question without the image or video, with the option to answer ``unsure''; a question was discarded if all three answered correctly. 

\section{Experiments and Results}
\label{sec:experiments_results}

\begin{figure*}[t]
    \centering
    \includegraphics[width=\linewidth]{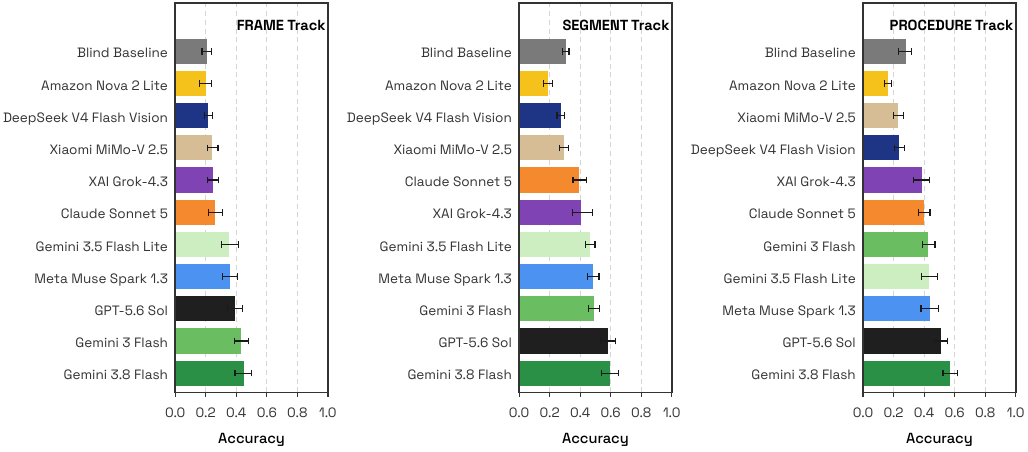}
    \caption{\textbf{Overall performance for all models and tracks.} Accuracy is the mean of per-video Accuracy. Uncertainty is depicted using bootstrapped 95\% \acp{ci}.}
    \label{fig:results_trackwise}
\end{figure*}

\paragraph{Metrics and ranking.} We evaluated model performance primarily using accuracy. Answers were parsed according to their expected format (Table~\ref{tab:scoring}). Open-ended, multiple-choice, and matching answers were graded for semantic equivalence by a \acl{llm} (\acs{llm}) judge (gpt-oss-120b, validated against two human raters in Appendix~\ref{app:judge}).

\paragraph{Inference and model selection.} All evaluations were conducted via cloud-based \acp{api} to ensure a standardized and reproducible evaluation environment. The benchmark includes a representative selection of current frontier \acp{vlm}, namely GPT-5.6 Sol, Claude Sonnet 5, Gemini 3.8 Flash, Gemini 3 Flash, Gemini 3.5 Flash Lite, Meta Muse Spark 1.3, xAI Grok-4.3, DeepSeek V4 Flash Vision, Amazon Nova 2 Lite, and Xiaomi MiMo-V 2.5. Details on the prompts provided to the \acp{vlm} can be found in Appendix~\ref{app:model_prompts}.

\paragraph{Data preprocessing.} The input pipeline was adapted to the native capabilities of each model. Depending on model support, inputs were provided either as native video or as chronologically ordered frame sequences. To ensure consistency across tracks while balancing temporal granularity and computational efficiency,  SEGMENT windows were sampled at 1 \ac{fps}, and PROCEDURE windows were represented by 300 uniformly spaced frames. Visual inputs were standardized to a spatial resolution of 210 $\times$ 360 pixels.

\paragraph{Baseline. }As baseline, we used a no-image setting in which GPT-5.6 Sol received only the textual prompt and question, but no visual input. This baseline assessed the extent to which questions can be answered from textual priors, dataset biases, or linguistic plausibility alone, without visual or temporal grounding. 

\paragraph{Results.} In the human baseline, three raters (two surgical residents and one medical student) answered 300 questions on the test videos, drawn from an earlier version of the question pool. They reached 62.0\%, 69.8\%, and 57.4\% Accuracy on the FRAME, SEGMENT, and PROCEDURE track, respectively (Appendix~\ref{app:human_baseline}).

The performance of the evaluated \acp{vlm} on the test data is summarized in Figure~\ref{fig:results_trackwise}. On the FRAME track, the highest overall Accuracy was obtained by Gemini 3.8 Flash (44.8\% [39.1, 50.1] 95\% \ac{ci}), followed by Gemini 3 Flash (43.2\% [39.0, 48.0]) and GPT-5.6 Sol (38.9\% [34.3, 44.0]), improving on the text-only baseline (20.7\%) by 24.1, 22.5, and 18.2\,\ac{pp}, respectively. Only half of the models clearly outperformed the baseline: Muse Spark 1.3 and Gemini 3.5 Flash Lite followed with gains of 15.0 and 14.3\,\ac{pp}, whereas the other five models (Claude Sonnet 5, Grok-4.3, MiMo-V 2.5, DeepSeek V4 Flash Vision, and Nova 2 Lite) stayed within 6\,\ac{pp} of it. \textit{Object recognition and identity matching} proved more challenging than \textit{aggregation} (see Figure~\ref{fig:results} in Appendix~\ref{app:further_results}), with \textit{object attributes} yielding the lowest scores (22.7\% averaged over all models; see Figure~\ref{fig:leaf_heatmap} in Appendix~\ref{app:further_results}).

Across the video tracks, \textit{event and procedural understanding} obtained the highest scores (average 56.5\% across all models), while none of the models achieved satisfactory performance on \textit{temporal grounding} tasks. With an average Accuracy of 41.7\% over all ten models, performance on the SEGMENT track exceeded the text-only baseline (30.4\%) by 11.3\,\ac{pp} on average. The highest overall Accuracy was obtained by Gemini 3.8 Flash (59.4\% [53.9, 64.9]), followed by GPT-5.6 Sol (58.3\% [53.2, 63.1]) and Gemini 3 Flash (49.1\% [45.3, 52.5]), improving on the baseline by 29.0, 27.9, and 18.7\,\ac{pp}, respectively. \textit{Temporal grounding} was the most difficult capability (see Figure~\ref{fig:results} in Appendix~\ref{app:further_results}), with a mean performance of 26.9\% averaged over all models.

On the PROCEDURE track, Accuracy averaged 37.9\% over the ten models (text-only baseline: 27.8\%). The highest overall Accuracy was obtained by Gemini 3.8 Flash (56.7\% [52.4, 62.0]), followed by GPT-5.6 Sol (51.0\% [46.7, 55.1]) and Muse Spark 1.3 (43.8\% [38.0, 49.3]). These three models improved on the baseline by 28.9, 23.2, and 16.0\,\ac{pp}, respectively, whereas DeepSeek V4 Flash Vision, MiMo-V 2.5, and Nova 2 Lite scored below it. \textit{Temporal grounding} dropped to 12.6\% averaged over all models, less than half of its SEGMENT value.

On the same 300 questions as the human raters, the best model on each track scored 12.0, 12.1, and 6.9\,\ac{pp} lower on the FRAME, SEGMENT, and PROCEDURE track, respectively (Table~\ref{tab:human} in Appendix~\ref{app:human_baseline}).

\section{Discussion}
\label{sec:discussion}

HeiCo-FOCUS uniquely evaluates long-context video understanding in a real-world, safety-critical setting. Unlike existing datasets that focus on short clips or isolated perception tasks, it requires models to maintain consistent representations of multiple objects over procedures lasting up to hours, where errors accumulate and cannot be corrected from local context. By anchoring evaluation in the concrete task of foreign object tracking and retrieval - an objective, verifiable, and clinically critical problem - the benchmark directly links technical performance to real-world utility. 

Importantly, our human baseline highlighted the difficulty of the long-context reasoning targeted by HeiCo-FOCUS. Even the human raters reached only 35\% on \textit{event aggregation} and 40\% on \textit{duration estimation} (20 questions each), both of which require following foreign objects over long time spans. This observation retrospectively supports our decision to establish persistent object identities through a dedicated, quality-controlled annotation process rather than asking clinicians to formulate and answer long-context questions directly from the videos. By grounding question generation in these curated instance trajectories, we reduce errors arising from ad-hoc manual reconstruction of object histories while retaining clinical expertise for question types that require higher-level medical reasoning.

To our knowledge, HeiCo-FOCUS is the first surgical \ac{vqa} benchmark to include questions spanning full procedures exceeding 20 minutes while explicitly evaluating persistent object tracking, temporal consistency, and long-horizon aggregation in surgery. However, constructing such a benchmark posed several challenges. First, defining consistent annotation rules for different foreign objects required careful clinical and technical consideration. For example, certain object types, such as surgical staples, can be too small and numerous to be reliably annotated individually by human annotators. Similarly, object-specific decisions, such as how to handle suture threads attached to needles, required detailed annotation guidelines, as documented in the appendix. Second, video artifacts, as well as limited temporal and spatial resolution, posed challenges even for expert annotators (see Appendix~\ref{app:challenges}). %
Third, designing a capability taxonomy that was simultaneously clinically meaningful, technically informative, and sufficiently disentangled for systematic evaluation required extensive iterative refinement between clinical and technical experts. We addressed many of these challenges through a rigorous transdisciplinary and quality-controlled annotation process involving crowd workers, medical students, surgeons, and surgical AI researchers. Nevertheless, we acknowledge that some annotation and taxonomy decisions may remain debatable. To promote transparency and continuous improvement, we therefore include a community feedback process that allows users to report potentially problematic or ambiguous cases. 

Several results deserve further discussion. Accuracy did not follow price: the most accurate model, Gemini 3.8 Flash, cost about a fifth of GPT-5.6 Sol per question on the video tracks (Figure~\ref{fig:cost_accuracy}).  While we captured statistical uncertainty via bootstrapping, we did not systematically investigate prompt sensitivity or variability across repeated model runs. Note that, in this context, evaluating long-context VLMs on extended surgical videos incurs substantial monetary and environmental costs (see Figure~\ref{fig:cost_accuracy}), highlighting the need to prioritize efficient evaluation protocols. Finally, the dataset was derived from a limited number of colorectal procedures, which may introduce biases related to surgical style, instrumentation, patient population, or recording conditions. Consequently, benchmark performance should not be interpreted as evidence of safe generalization across procedures, institutions, or clinical settings. Future work should expand the diversity of procedures, institutions, and surgical settings represented in the benchmark, while placing particular emphasis on increasing the number and diversity of expert-generated reasoning questions. 

In conclusion, through its clinically grounded design, capability-centric evaluation, and rigorous multi-stage annotation process, HeiCo-FOCUS establishes a fundamentally new testbed for assessing whether modern \acp{vlm} can move beyond short-term pattern recognition toward reliable long-horizon reasoning. By exposing failures in cumulative temporal consistency, it provides a foundation for developing and evaluating models that can operate reliably in complex, real-world settings.

\section*{Acknowledgments}
B.G. is funded by the German Research Foundation (DFG, Deutsche Forschungsgemeinschaft) as part of Reinhart Koselleck-project -- Project ID 560101272 -- SIMSURGE: Balancing the odds by simulating rare cases for surgical data science.
A.H.'s work is partly supported by the Federal Ministry of Research, Technology and Space in DAAD project 57616814 (SECAI, School of Embedded Composite AI, \url{https://secai.org/}) as part of the program Konrad Zuse Schools of Excellence in Artificial Intelligence.
T.R. was supported by a scholarship from the Hanns Seidel Foundation, funded by the German Federal Ministry of Research, Technology and Space (BMFTR).
F.R.K. receives support from the German Cancer Research Center (CoBot 2.0), the Joachim Herz Foundation (Add-On Fellowship for Interdisciplinary Life Science), the Central Indiana Corporate Partnership AnalytiXIN Initiative, the Evan and Sue Ann Werling Pancreatic Cancer Research Fund, and the Indiana Clinical and Translational Sciences Institute (EPAR4157) funded, in part, by Grant Number UM1TR004402 from the National Institutes of Health, National Center for Advancing Translational Sciences, Clinical and Translational Sciences Award. The content is solely the responsibility of the authors and does not necessarily represent the official views of the National Institutes of Health.

This work was partly supported by the German Research Foundation (DFG, Deutsche Forschungsgemeinschaft) as part of Germany's Excellence Strategy -- EXC 2050/2 -- Project ID 390696704 -- Cluster of Excellence ``Centre for Tactile Internet with Human-in-the-Loop'' (CeTI) of TUD; the Federal Ministry of Research, Technology, and Space (BMFTR) as part of the research program Communication Systems ``Souverän. Digital. Vernetzt.'' -- joint project 6G-life with project ID 16KIS2413K; project ``Next Generation AI Computing (gAIn)'' funded by the Bavarian Ministry of Science and the Arts and the Saxon Ministry for Science, Culture, and Tourism; the BMFTR in DAAD project 57616814 (SECAI, \url{https://secai.org/}) and the DFG as part of Reinhart Koselleck-project -- Project ID 560101272 -- SIMSURGE.

\section*{Disclosures}
T.W. was the former Program Director for the Wellcome Leap SAVE program. The funder had no additional input into the decision to write or submit this work for publication.
A.R. discloses support for the research of this work from the Helmholtz Association of German Research Centers in the scope of the Helmholtz Imaging Incubator (HI).

\bibliographystyle{iclr2027_conference} 
\bibliography{references}

\appendix
\clearpage

\section{Comparison with Existing Surgical \ac{vqa} Datasets}
\label{app:dataset_comparison}

\begin{table*}[ht]
\centering
\small
\renewcommand{\arraystretch}{1.0}
\caption{Overview of publicly released surgical \acf{vqa} datasets. Class labels refers to \acp{vqa} generated from classification labels extracted from available vision datasets.
*Full dataset not released yet.
}
\label{tab:surgical_vqa_comparison}

\resizebox{\textwidth}{!}{
\begin{tabular}{lccccccc}
\toprule
\textbf{Dataset} &
\textbf{\#Videos} &
\textbf{\#VQAs} &
\multicolumn{3}{c}{\textbf{Context Length}} &
\textbf{Source}  \\
\cmidrule(lr){4-6}
& & & \textbf{Frame} & \textbf{Segment} & \textbf{Procedure} & & \\
& & &  & \textbf{(30 sec$\leq$ t$\leq$5 min)} & \textbf{  t$>$5 min } & & \\
\midrule

Cholec80-VQA~\citep{seenivasan2022surgical} & 40 & 43,182 & 43,182 & 0 & 0 & Class. labels   \\
EndoVis-18-VQA~\citep{seenivasan2022surgical} & 14 & 11,783 & 11,783 & 0 & 0 & Class. labels  \\
EndoVis-VQLA~\citep{bai2023surgical} & 24 & 12,255 & 12,255 & 0 & 0 & Class. labels  \\
PSI-AVA-VQA~\citep{seenivasan2023surgicalgpt} & 8 & 10,291 & 10,291 & 0 & 0 & Class. labels  \\
PiTVQA~\citep{he2024pitvqa} & 25 & 884,242 & 884,242 & 0 & 0 & Class. labels  \\
SSG-VQA~\citep{yuan2024advancing} & 50 & 960,000 & 960,000 & 0 & 0 & Class. labels  \\
Surg-396K~\citep{wang2025endochat} & 109 & 396,000 & 396,000 & 0 & 0 & Class. labels  \\
SurgVLM-DB~\citep{zeng2025surgvlm} & 849 & 7,798,400 & 7,798,400 & 0 & 0 & Class. labels  \\
SurgPub-Video~\citep{li2026surgpub} & 3,538 & 48,520 & 0 & 48,520 & 0 & Audio transcripts  \\
Surg-LaVI~\citep{perez2026surglavi} & 5,317 & 240,000 & 77$^{*}$ & 117,061$^{*}$ & 320$^{*}$ & Audio transcripts \\
\midrule
\textbf{Ours} & \textbf{30} & \textbf{30,000} & \textbf{12,000} & \textbf{12,000} & \textbf{6,000} & \textbf{Expert annotations}  \\
\bottomrule
\end{tabular}
}
\end{table*}

\section{Data Access and Usage}
\label{app:data_access_usage}

The dataset will be hosted through Hugging Face datasets \citep{Lhoest_Datasets_A_Community_2021} and published there soon; until then, the corresponding authors provide access on request.
We stay compliant with the datasets \ac{api} and a single call allows access to all question answer pairs. For seamless integration with the loading and processing of videos for each of the tracks, we also publish a Python utilities package through PyPI and open source the code at \url{https://github.com/IMSY-DKFZ/orena-focus}.
The usage of the package is shown in Figure \ref{lst:usage}, with more involved data loading classes and additional examples provided in the repository.

\begin{figure}[bh!]
\begin{lstlisting}
from focus import (FocusDataset, DatasetSplit, Track,
                   FocusConfig, set_config, download,
                   Evaluator, Response)

# configure local data root and download video files
set_config(FocusConfig(root_dir="/data/focus"))
download("heico")

# QA annotations are fetched automatically from HuggingFace
ds = FocusDataset("heico", DatasetSplit.TEST, Track.SEGMENT)

responses = [
    Response(qID=req.qID, content=model(req))
    for req, _ in ds
]

results, summary = Evaluator().run(
    requests=ds.requests,
    references=ds.references,
    responses=responses,
)
\end{lstlisting}
\caption{Loading and evaluating a model on \mbox{HeiCo-FOCUS} using the \texttt{orena-focus} Python package. Video files are downloaded once via \texttt{download()}, and question--answer annotations are fetched automatically from Hugging Face. The \texttt{Evaluator} handles answer-format parsing and routes open-ended and multiple-choice answers to an \ac{llm} judge.}
\label{lst:usage}
\end{figure}

\section{Taxonomy}
\label{app:taxonomy}

\textbf{Object recognition and identity matching}: Recognition of object instances, their semantic type, attributes, spatial context, and identity consistency across time. Subcategories:
\begin{itemize}
    \item Object identification: semantic classification of visible object instances
    \item Object instance identity matching: correspondence of object instances across time
    \item Object attributes and state: recognition of observable object properties or handling states
    \item Object spatial localization (camera): object localization relative to the image plane
    \item Object spatial localization (situs): object localization relative to anatomical structures
\end{itemize}

\textbf{Temporal grounding}: Localization of events or object occurrences within the video timeline, including their temporal position and duration. Subcategories:

\begin{itemize}
    \item Temporal localization: identification of the time point or interval at which an event or object occurrence takes place
    \item Duration estimation: estimation of the temporal extent for which an event or object occurrence persists
\end{itemize}

\textbf{Aggregation}: Combination of information on foreign objects and events across multiple objects, instances, and/or time points. Subcategories:

\begin{itemize}
    \item Object aggregation: aggregation of information across multiple foreign object instances and/or categories
    \item Event aggregation: aggregation of events that involve foreign objects across time
\end{itemize}

\textbf{Event and procedural understanding}: Recognition and interpretation of actions, events, and their procedural structure over time. Subcategories:

\begin{itemize}
    \item Foreign object interaction recognition: identification of actions or manipulations involving a foreign object
    \item Foreign object usage purpose: identification of the general procedural role of a foreign object at a given time. Answers are grounded in what is visually demonstrated; no inference about hypothetical outcomes or hidden intent
    \item Temporal ordering: determination of the relative order of events or procedural steps, including the identification of the n-th event in a series of events
\end{itemize}

\textbf{Complex reasoning}: Inference of functional, causal, relational, or outcome-related information beyond direct observation, and reasoning over interactions, dependencies, comparative patterns, and temporally extended evidence across objects and events. Subcategories:
\begin{itemize}
    \item Functional reasoning: inference of the intended function or role of an object or action beyond what is directly shown
    \item Causal and consequence reasoning: inference of cause–effect relationships between actions, events, or object states and inference of potential or actual outcomes resulting from actions or events
    \item Multi-step compositional reasoning: inference that requires the execution of a sequence of interdependent reasoning steps across objects, space, and time, involving the decomposition of a query into multiple sub-tasks to derive an answer that cannot be attributed to a single dominant capability
\end{itemize}
See Figure~\ref{fig:taxonomy} for examples.

\newpage
\section{VQA Generation Process}
\label{app:vqa_generation_process}

\begin{longtable}{
>{\raggedright\arraybackslash}p{1.8cm}
>{\raggedright\arraybackslash}p{1.9cm}
>{\raggedright\arraybackslash}p{1.9cm}
>{\raggedright\arraybackslash}p{1.9cm}
>{\raggedright\arraybackslash}p{1.9cm}
>{\raggedright\arraybackslash}p{1.9cm}
}

\caption{Overview of the six-stage annotation pipeline used to construct the HeiCo-FOCUS benchmark. Raw surgical videos are progressively transformed into clinically grounded \acf{vqa} pairs through foreign object screening, frame-level localization, automated refinement, instance-level tracking with anchor moment identification, and multi-stage question generation and verification. The table summarizes inputs, outputs, annotation scale across 30 videos, and the combination of crowd, expert, and automated methods used at each stage.}
\label{tab:annotation_stages} \\

\toprule
\textbf{Stage} & \textbf{Purpose} & \textbf{Input} & \textbf{Output} & \textbf{\#annotations (total across 30 videos)} & \textbf{Annotators / Method} \\
\midrule
\endfirsthead

\toprule
\textbf{Stage} & \textbf{Purpose} & \textbf{Input} & \textbf{Output} & \textbf{\#annotations (total across 30 videos)} & \textbf{Annotators / Method} \\
\midrule
\endhead

\endlastfoot

\textbf{Stage 1} Foreign object presence screening 
& Identification of video segments with foreign objects 
& 5 s long videos 
& Presence flag for each class of foreign objects on each clip
& 69,469 annotated 5 s videos; 16\% contained a foreign object
& Quality Match GmbH, Häusserstraße 36, 69115 Heidelberg, Germany \\

\midrule

\textbf{Stage 2} Frame-level localization (bounding boxes at 1 fps) 
& Identification and localization of foreign objects in individual frames 
& Frames (sampled at 1 fps) corresponding to clips with foreign objects 
& Frames (sampled at 1 fps) with bounding boxes and class assignment for all foreign objects 
& 53,001 bounding boxes on a total of 54,528 frames 
& Quality Match GmbH, Häusserstraße 36, 69115 Heidelberg, Germany; iMerit, 160 West Santa Clara St., Suite 600, San Jose, CA 95113, USA \\

\midrule

\textbf{Stage 3} Automated bounding box completion / correction
& Automatic refinement of crowd-based foreign object identification and localization
& Frames (sampled at 1 fps) with bounding boxes and class assignment for all foreign objects
& Automatically corrected frames (sampled at 1 fps) with bounding boxes and class assignment for all foreign objects
& 4,832 corrected frames
& Automated method \\

\midrule

\textbf{Stage 4} Instance assignment
& Instance assignment to individual objects (clips only first three frames), final correction of bounding boxes, and class assignments.
& Frames (sampled at 1 fps) with bounding boxes and class assignment for all foreign objects
& Frames (sampled at 1 fps) with class and instance assignments and bounding boxes around all foreign objects
& 79,682 bounding boxes (instance identities for all classes except clips); 324,273 curated frames
& 18 Medical students \\

\midrule

\textbf{Stage 5} VQA generation and clinical question design
& Generation of questions and answers based on foreign object information (auto-generated + expert-added)
& Frames (sampled at 1 fps) with class and instance assignments, bounding boxes around all foreign objects, and anchor moments
& VQA pairs balanced across the 5 core capabilities, including 15 sub-capabilities
& 30,000 VQA pairs balanced across the 5 capability groups.\newline\newline
FRAME track: 12,000 VQA pairs\newline
SEGMENT track: 12,000 VQA pairs\newline
PROCEDURE track: 6,000 VQA pairs
& 18 Medical students and 7 expert surgeons;\newline\newline
automatic method based on instance information and expert annotations \\

\midrule

\textbf{Stage 6} End-to-end VQA verification
& Verification of the question generation pipeline 
& A spreadsheet (via link) containing question IDs, corresponding timestamps, and the associated surgical procedure video
& The same spreadsheet completed with answers, a confidence score (1–10), optional comments, and, where necessary, a “Needs revision” flag
& 500 verified questions
& 14 Surgical AI researchers + 3 Medical students + 5 expert surgeons \\

\bottomrule

\end{longtable}

Labeling instructions were provided separately for all stages that required human involvement.

\begin{figure}[h]
    \centering
    \includegraphics[width=0.93\linewidth]{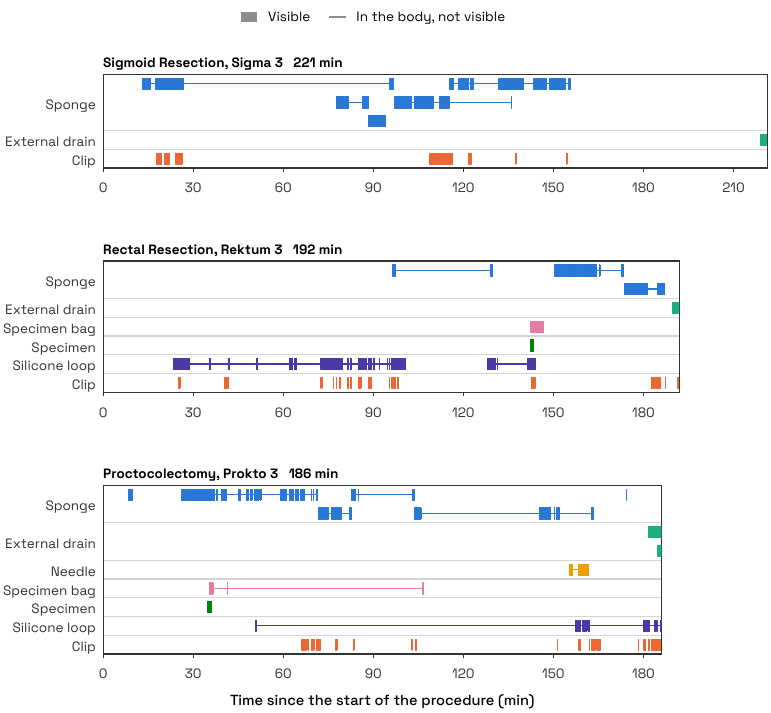}
    \caption{\textbf{Foreign-object timelines from Stage 4.} One video per procedure type on a common time axis. Bars mark the seconds in which an instance is visible; thin lines join its first and last sighting. Instances of one class share a row when they are at least three minutes apart. Clips carry no instance identities and form one visibility row.}
    \label{fig:fo_timeline}
\end{figure}

Figure~\ref{fig:fo_timeline} shows the Stage 4 annotations of one video per procedure type.

\section{Model Prompts}
\label{app:model_prompts}

We provide details on the prompt design used for all experiments. Each prompt is composed of four components: (i) a context prefix describing the surgical setting, (ii) a foreign object definition (see Appendix~\ref{subapp:foreign_object_defs}), (iii) the task-specific question, and (iv) format specifications that enforce structured outputs. Appendix~\ref{app:example_prompts} shows a complete prompt for one SEGMENT question.

\subsection{Context Prefix}

\textbf{FRAME}

\begin{quote}
"You are assisting in the analysis of laparoscopic videos from minimally invasive surgeries. Specifically, you see a frame from a \{procedure\_type\} at timepoint \{timestamp\}. Analyze the footage and answer the question based on the visual evidence. Be as concise as possible by following the output format instructions."
\end{quote}

\textbf{SEGMENT}

\begin{quote}
"You are assisting in the analysis of laparoscopic videos from minimally invasive surgeries. Specifically, you see a video from a \{procedure\_type\} from timepoint \{timestamp\_start\} to \{timestamp\_end\}. The video is sampled at 1 fps. Analyze the footage and answer the question based on the visual evidence. Be precise and concise."
\end{quote}

\textbf{PROCEDURE}

\begin{quote}
"You are assisting in the analysis of laparoscopic videos from minimally invasive surgeries. Specifically, you see a video from a \{procedure\_type\} from timepoint \{timestamp\_start\} to \{timestamp\_end\}. Analyze the footage and answer the question based on the visual evidence. Be precise and concise."
\end{quote}

\subsection{Foreign Object Definitions}
\label{subapp:foreign_object_defs}

\texttt{=== FOREIGN OBJECT DEFINITION ===}

A foreign object is any object fully introduced into the patient's body cavity during surgery that must be retrieved or accounted for. Importantly, standard surgical instruments that remain connected to the external environment (e.g., graspers, scissors, trocars, staplers, cameras) are not considered foreign objects. Furthermore, we exclude detachable parts of surgical instruments, particularly anvil components of staplers.

\texttt{=== FOREIGN OBJECT CLASSES ===}

The following are the only classes to consider. Do not infer or name any class outside this list, such as staples.

\begin{itemize}
    \item \textbf{Sponge}: A soft, absorbent material used to soak up fluids. They are typically white when fresh and can become reddish-brown when saturated with blood.

    \item \textbf{External drain}: A clear or fluid-filled tube used to evacuate fluids from the surgical site to the outside of the body. Its tip may be temporarily visible within the surgical field during placement or adjustment.

    \item \textbf{Needle}: A sharp, pointed metal instrument, straight or curved, used for placing sutures. A needle is considered visible only if the metallic needle itself is seen. The suture thread alone does not qualify as needle visibility.

    \item \textbf{Clip}: A small metal or polymer device used to seal vessels or ducts. May potentially remain in the body. Clips only count as foreign objects once placed in the abdomen. They do not count as foreign objects while loaded within the clip applier instrument.

    \item \textbf{Specimen bag}: A sterile pouch used to collect and retrieve resected tissue or organs from the body cavity. Only consider the pouch itself as a foreign object and ignore the string attached to it.

    \item \textbf{Silicone loop}: A soft and flexible, typically white band used to encircle and control blood vessels or structures for isolation and traction.

    \item \textbf{Specimen}: Excised biological tissue (e.g., appendix, resected bowel segment) that must be retrieved from the body cavity before completion of the procedure. Tissue is considered a foreign object once it is fully detached from surrounding anatomy. This excludes fat and blood. Specimens contained within a specimen bag are considered retrieved; specimen bags are annotated separately.

    \item \textbf{Absorbable hemostatic agent}: A soft, absorbable material (for example oxidized cellulose or a gelatin sponge) applied to a bleeding surface to promote clotting. It may be intentionally left inside the body cavity.
\end{itemize}

\texttt{=== TEMPORAL AND VISIBILITY RULES ===}

\begin{itemize}
    \item A foreign object is considered inserted / introduced at the moment it first becomes partially visible inside the body cavity. In case of clips, it is also required to leave the clip applier. Class-specific visibility exceptions (e.g., needle string) take precedence.

    \item A foreign object is considered present as long as it is inside the body cavity, even during periods of occlusion or when temporarily off-screen.

    \item A foreign object is considered removed / retrieved at the moment it fully exits the body cavity or is fully inserted into a specimen bag. A specimen fully inside a specimen bag is considered retrieved; the specimen bag itself remains present until it exits the body cavity.

    \item If the same physical object reappears after occlusion, it is the same instance — do not count it again. When multiple instances of the same class are present, reappearing objects should be attributed to already-known instances before assuming a new one.

    \item All time references are given relative to the start of the intervention (00:00:00).
\end{itemize}

\subsection{Format Instructions}

The following output type-specific instructions are added to the prompt:

\begin{itemize}
    \item \textbf{Time point}: Return the answer only in hh:mm:ss format without any additional text and with respect to the start of the procedure (not necessarily the start of the video clip). Example: 01:37:12
    
    \item \textbf{Duration}: Return the answer only in hh:mm:ss format without any additional text. Example: 00:02:13
    
    \item \textbf{Foreign object class}: Return only labels from the foreign objects list. Example: Sponge
    
    \item \textbf{Number}: Return only a single integer value. Example: 3

    \item \textbf{Percentage}: Return only a single percentage value between 0 and 100 without the percent sign. Example: 75
    
    \item \textbf{Binary}: Return only yes or no. Example: yes
    
    \item \textbf{Multiple Choice}: Return only one of the provided options.
    
    \item \textbf{Open Ended}: Answer the question directly using only the most relevant details. Do not add lengthy explanations or rephrasings of the question.
    
    \item \textbf{All (postfix)}: Strictly adhere to these output format specifications. Do not add any extra text.
\end{itemize}

\subsection{Scoring}
\label{app:scoring}
Each answer is parsed according to its expected format and scored with the rule in Table~\ref{tab:scoring}.

\begin{table}[h]
\centering
\small
\caption{Scoring rule per answer format. $d$ is the length of the input window in seconds. Answers that do not parse in the expected format count as incorrect.}
\label{tab:scoring}
\begin{tabular}{lp{8cm}}
\toprule
Format & Counted as correct if \\
\midrule
Number & the integer equals the reference \\
Yes/no & the answer equals the reference (case-insensitive) \\
Foreign-object class & the set of named classes equals the reference set (order- and case-insensitive; ``none'' only on its own) \\
Percentage & within $\pm$5\,\ac{pp} of the reference \\
Time point, duration & within $\tau=\min(5,\,1+4d/360)$\,s of the reference: 1.3, 2.3, and 4.3\,s for 30\,s, 2\,min, and 5\,min windows; 5\,s for windows of 6\,min or longer \\
Open-ended, multiple choice, matching & judged semantically equivalent to the reference by gpt-oss-120b; answers longer than 300 characters count as incorrect \\
\bottomrule
\end{tabular}
\end{table}

\subsection{Example Prompt}
\label{app:example_prompts}

\begin{quote}
\ttfamily

You are assisting in the analysis of laparoscopic videos from minimally invasive surgeries. Specifically, you see a video from a Sigmoid Resection from timepoint \texttt{00:14:40} to \texttt{00:19:21}. The video is sampled at 1 fps.

\vspace{0.5em}

Analyze the footage and answer the question based on the visual evidence. Be precise and concise.

\vspace{0.5em}

\texttt{=== FOREIGN OBJECT DEFINITION ===}

A foreign object is any object fully introduced into the patient’s body cavity during surgery that must be retrieved or accounted for. Importantly, standard surgical instruments that remain connected to the external environment (e.g., graspers, scissors, trocars, staplers, cameras) are not considered foreign objects. Furthermore, we exclude detachable parts of surgical instruments, particularly anvil components of staplers.

\texttt{=== FOREIGN OBJECT CLASSES ===}

The following are the only classes to consider. Do not infer or name any class outside this list, such as staples.

Sponge: A soft, absorbent material used to soak up fluids. They are typically white when fresh and can become reddish-brown when saturated with blood.

External drain: A clear or fluid-filled tube used to evacuate fluids from the surgical site to the outside of the body. Its tip may be temporarily visible within the surgical field during placement or adjustment.

Needle: A sharp, pointed metal instrument, straight or curved, used for placing sutures. A needle is considered visible only if the metallic needle itself is seen. The suture thread alone does not qualify as needle visibility.

Clip: A small metal or polymer device used to seal vessels or ducts. May potentially remain in the body. Clips only count as foreign objects once placed in the abdomen. They do not count as foreign objects while loaded within the clip applier instrument.

Specimen bag: A sterile pouch used to collect and retrieve resected tissue or organs from the body cavity. Only consider the pouch itself as foreign object and ignore the string attached to it.

Silicone loop: A soft and flexible, typically white band used to encircle and control blood vessels or structures for isolation and traction.

Specimen: Excised biological tissue (e.g., appendix, resected bowel segment) that must be retrieved from the body cavity before completion of the procedure. Tissue is considered a foreign object once it is fully detached from surrounding anatomy. This excludes fat and blood. Specimens contained within a specimen bag are considered retrieved; specimen bags are annotated separately.

Absorbable hemostatic agent: A soft, absorbable material (for example oxidized cellulose or a gelatin sponge) applied to a bleeding surface to promote clotting. It may be intentionally left inside the body cavity.

\texttt{=== TEMPORAL AND VISIBILITY RULES ===}

A foreign object is considered inserted / introduced at the moment it first becomes partially visible inside the body cavity. In case of clips, it is also required to leave the clip applier. Class-specific visibility exceptions (e.g., needle string) take precedence.

A foreign object is considered present as long as it is inside the body cavity, even during periods of occlusion or when temporarily off-screen.

A foreign object is considered removed / retrieved at the moment it fully exits the body cavity or is fully inserted into a specimen bag. A specimen fully inside a specimen bag is considered retrieved; the specimen bag itself remains present until it exits the body cavity.

If the same physical object reappears after occlusion, it is the same instance -- do not count it again. When multiple instances of the same class are present, reappearing objects should be attributed to already-known instances before assuming a new one.

All time references are given relative to the start of the intervention (\texttt{00:00:00}).

\vspace{0.5em}

After the Sponge was inserted in the abdomen, which different foreign object classes were inserted or created (in case of a specimen) afterwards, if any?

\vspace{0.5em}

Please provide the class name(s) or answer none. Return only labels from the foreign objects list above. Example: Sponge. Strictly adhere to these output format specifications. Do not add any extra text.

\end{quote}

\section{Validation of the \ac{llm} judge}
\label{app:judge}
Open-ended, multiple-choice, and matching answers are graded by gpt-oss-120b. To test the judge against human grading, we drew 164 open-ended answers (76 FRAME, 76 SEGMENT, and 12 PROCEDURE), stratified by track and by the verdict of a judge model (Llama 4 Maverick), with equal numbers of accepted and rejected answers. Two raters, a surgical resident and a surgical data science researcher, graded each answer independently. Like the judge, they did not see the judge's verdict. Weighted to the population the answers were drawn from, gpt-oss-120b agrees with the resident on 90.7\% of answers ($\kappa = 0.74$) and with the second rater on 92.7\% ($\kappa = 0.78$), while the two raters agree with each other on 91.8\% ($\kappa = 0.78$). The judge is the stricter grader in 13 of its 15 disagreements with the resident and in 7 of 13 with the second rater, so grading by the raters would, if anything, raise the reported accuracies.

\section{Human Baseline}
\label{app:human_baseline}

\begin{table}[h]
\centering
\small
\caption{Accuracy (\%) of human raters and models on the same questions. Three raters (two surgical residents and one medical student) answered 300 test questions in total. Their answers were graded identically to the models.}
\label{tab:human}
\begin{tabular}{lrrr}
\toprule
 & FRAME & SEGMENT & PROCEDURE \\
\midrule
Human raters & \textbf{62.0} & \textbf{69.8} & \textbf{57.4} \\
Gemini 3.8 Flash         & 46.0 & 57.7 & 50.5 \\
GPT-5.6 Sol              & 40.0 & 55.0 & 45.5 \\
Gemini 3 Flash           & 50.0 & 49.0 & 39.6 \\
Meta Muse Spark 1.3      & 30.0 & 49.0 & 42.6 \\
Gemini 3.5 Flash Lite    & 44.0 & 46.3 & 42.6 \\
xAI Grok-4.3        & 28.0 & 38.9 & 33.7 \\
Claude Sonnet 5          & 38.0 & 38.3 & 29.7 \\
DeepSeek V4 Flash Vision & 34.0 & 28.9 & 16.8 \\
Xiaomi MiMo-V 2.5        & 24.0 & 28.2 & 18.8 \\
Amazon Nova 2 Lite       & 12.0 & 26.2 & 15.8 \\
Text-only baseline       & 24.0 & 35.6 & 25.7 \\

\bottomrule
\end{tabular}
\end{table}

\newpage
\section{Further Results}
\label{app:further_results}

\begin{figure}[h]
    \centering
    \includegraphics[width=\linewidth]{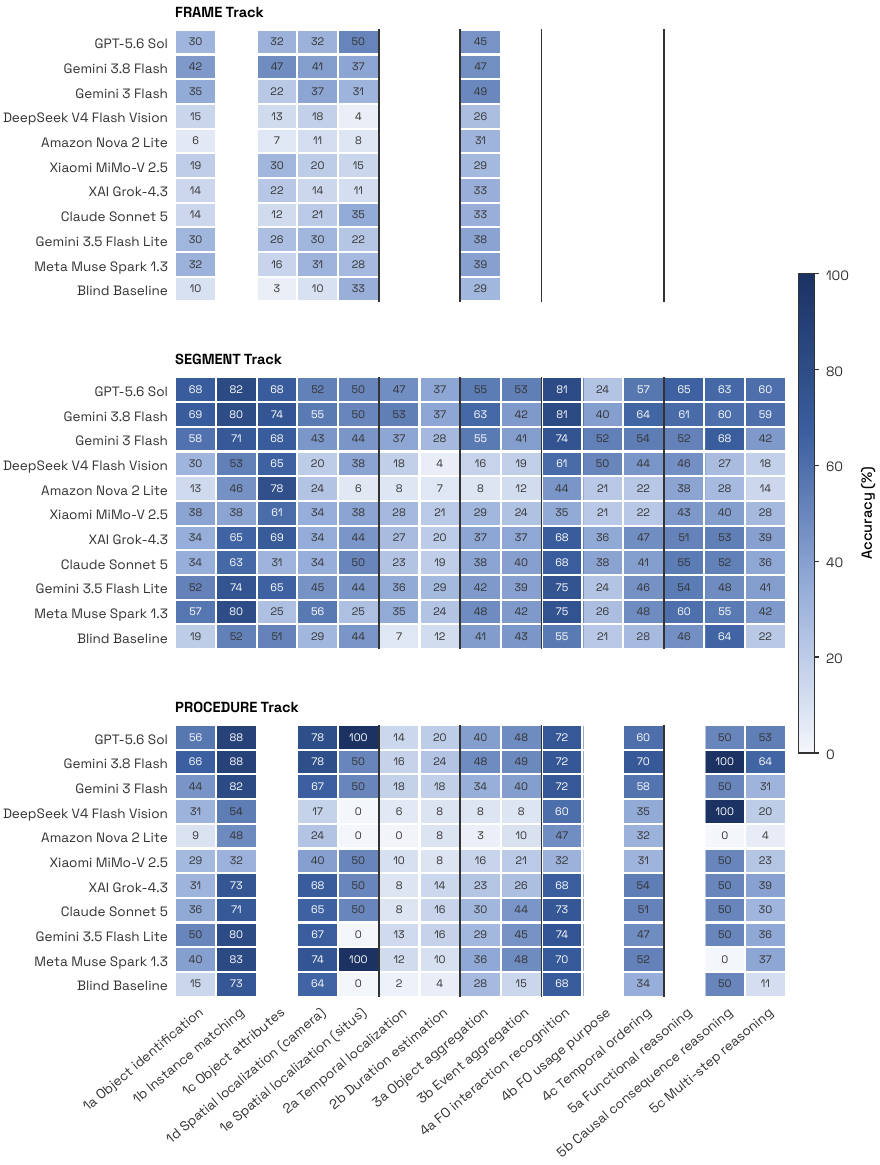}
    \caption{\textbf{Accuracy per sub-capability.} Mean of per-video Accuracy (\%) for each model and sub-capability; blank cells have no questions of that sub-capability in the track. Blind Baseline is the text-only baseline.}
    \label{fig:leaf_heatmap}
\end{figure}

\begin{figure}[h]
    \centering
    \includegraphics[width=0.88\linewidth]{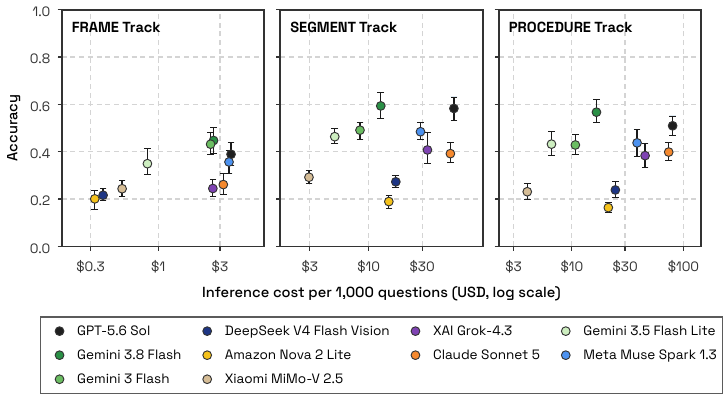}
    \caption{\textbf{Accuracy and inference cost.} Accuracy with 95\% \acp{ci}, as in Figure~\ref{fig:results_trackwise}, against the API cost per 1,000 test questions (log scale), for the models that see the video. Cost is the sum of the API charges for the answers, without judge calls.}
    \label{fig:cost_accuracy}
\end{figure}

Figure~\ref{fig:leaf_heatmap} gives Accuracy per sub-capability, and Figure~\ref{fig:cost_accuracy} plots Accuracy against the API cost of each run.

\begin{figure*}[h]
    \centering
    \includegraphics[width=0.8\linewidth]{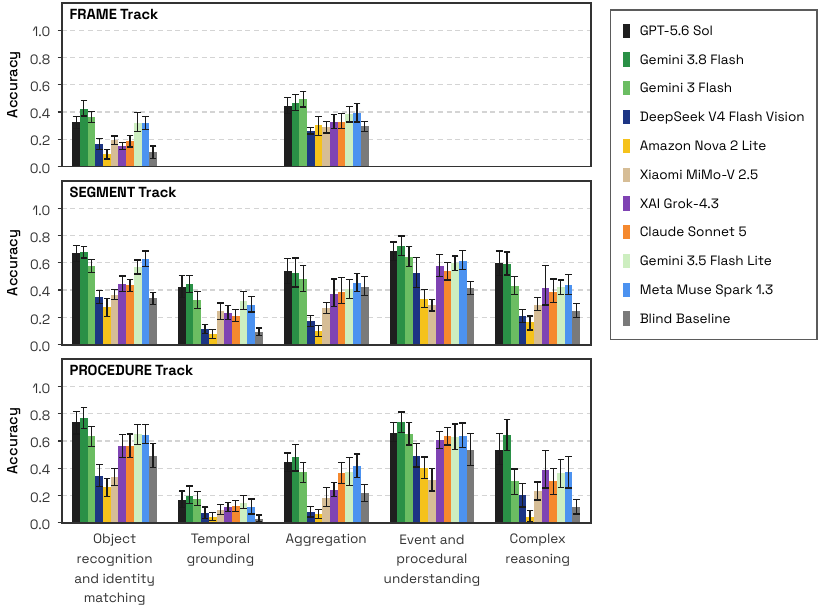}
    \caption{\textbf{Core results of the FRAME, SEGMENT, and PROCEDURE tracks.}
Performance of models stratified by capability group. Points are means of per-video Accuracy over the test videos; 95\% \acp{ci} from a two-level bootstrap over videos and questions (1,000 resamples). }
    \label{fig:results}
\end{figure*}

\clearpage %
\section{Challenging Examples}
\label{app:challenges}

\begin{longtable}{
>{\raggedright\arraybackslash}m{6.6cm}
>{\raggedright\arraybackslash}m{6.6cm}
}

\caption{Sample images illustrating some of the annotation challenges the annotators faced when generating the instance annotations as the basis for the \acf{vqa} generation pipeline.}
\label{tab:challenging_examples} \\

\toprule
\textbf{Annotation issue/challenge} & \textbf{Sample image} \\
\midrule
\endfirsthead

\toprule
\textbf{Annotation issue/challenge} & \textbf{Sample image} \\
\midrule
\endhead

\endlastfoot

\parbox[t]{\linewidth}{
Sponges tucked deep into the surgical site are challenging to identify and link.
}
&
\parbox[t]{\linewidth}{
\includegraphics[width=\linewidth]{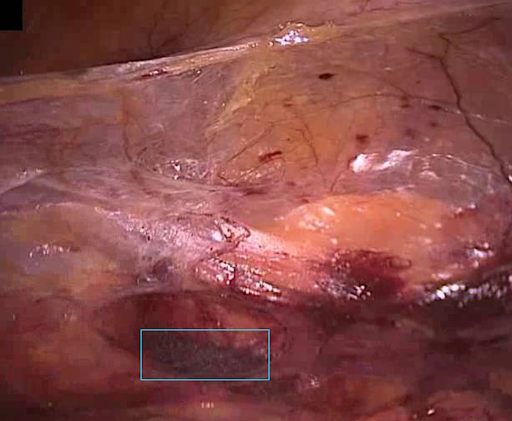}
}
\\

\parbox[t]{\linewidth}{
}
&
\parbox[t]{\linewidth}{
\includegraphics[width=\linewidth]{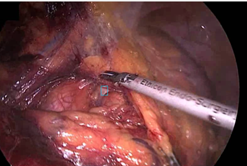}
}
\\

\midrule

\parbox[t]{\linewidth}{
Staples can be confused with clips and are infeasible to count.
}
&
\parbox[t]{\linewidth}{
\includegraphics[width=\linewidth]{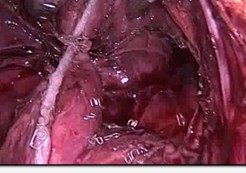}
}
\\

\midrule

\parbox[t]{\linewidth}{
Specimen and clips attached to them should not be annotated as foreign objects as soon as they are placed in the specimen bag.

}
&
\parbox[t]{\linewidth}{
\includegraphics[width=\linewidth]{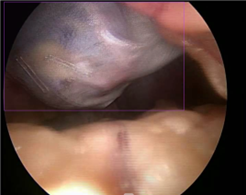}
}
\\

\midrule

\parbox[t]{\linewidth}{
Due to inadequate lighting at the image borders, distinguishing between shadows and blood-soaked sponges is challenging.

}
&
\parbox[t]{\linewidth}{
\includegraphics[width=\linewidth]{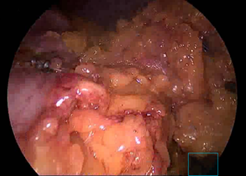}
}
\\

\midrule

\parbox[t]{\linewidth}{
For several sponges, it was difficult to decide whether they had been removed from the surgical site or remained in situ only to reappear later, particularly when the video was interrupted by a prolonged blue screen. }
&
\parbox[t]{\linewidth}{
\includegraphics[width=\linewidth]{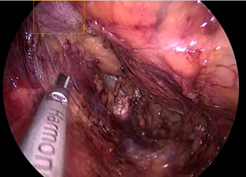}
}
\\

\parbox[t]{\linewidth}{
}
&
\parbox[t]{\linewidth}{
\includegraphics[width=\linewidth]{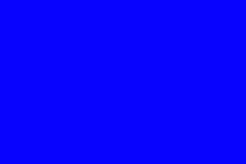}
}
\\

\midrule

\parbox[t]{\linewidth}{
Due to the dense local arrangement of clips, perspective shifts sometimes led to ambiguities in identity association, making it challenging to distinguish one clip from another. 

}
&
\parbox[t]{\linewidth}{
\includegraphics[width=\linewidth]{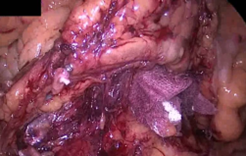}
}
\\

\parbox[t]{\linewidth}{
}
&
\parbox[t]{\linewidth}{
\includegraphics[width=\linewidth]{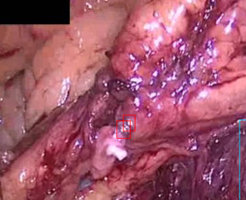}
}
\\

\midrule

\bottomrule

\end{longtable}

\end{document}